\documentclass[conference,a4paper]{IEEEtran}
\IEEEoverridecommandlockouts

\usepackage{xr-hyper} 
\usepackage{hyperref} 
\usepackage[acronym, nonumberlist, nogroupskip, style=super]{glossaries}
\usepackage{hyperref}
\newacronym{iot}{IoT}{Internet of Things} 

\usepackage{cite}
\usepackage{amsmath, amssymb, amsfonts}
\usepackage{graphicx}
\usepackage{textcomp}
\usepackage{bmpsize}
\usepackage[dvipsnames]{xcolor}
\usepackage{lipsum}

\usepackage{multirow}
\usepackage{tabularx}

\usepackage[T1]{fontenc}

\usepackage{subfig}	
\usepackage{soul}
\usepackage{booktabs}

\usepackage{multirow}
\usepackage{rotating}
\usepackage{booktabs}

\usepackage{color, colortbl}

\usepackage{comment}

\usepackage{cite}
\usepackage{amsmath, amssymb, amsfonts}
\usepackage{algorithmic}
\usepackage{graphicx}
\usepackage{textcomp}
\usepackage{xcolor}

\def\BibTeX{{\rm B\kern-. 05em{\sc i\kern-. 025em b}\kern-. 08em
 T\kern-. 1667em\lower. 7ex\hbox{E}\kern-. 125emX}}
\begin{document}

\title{WiFi Based Distance Estimation Using Supervised Machine Learning\
	\thanks{}
}

\author{
    \IEEEauthorblockN{Kahraman Kostas\IEEEauthorrefmark{1}\IEEEauthorrefmark{2}, Rabia Yasa Kostas\IEEEauthorrefmark{1}\IEEEauthorrefmark{3}, Francisco Zampella\IEEEauthorrefmark{1}, Firas Alsehly\IEEEauthorrefmark{1}}
    \IEEEauthorblockA{\IEEEauthorrefmark{1}Edinburgh Research Center, Huawei Technologies Co., Ltd. United Kingdom}
    \IEEEauthorblockA{\IEEEauthorrefmark{2}School of Mathematical \& Computer Sciences, Heriot Watt University, Edinburgh, United Kingdom}
    \IEEEauthorblockA{\IEEEauthorrefmark{3}School of Informatics, University of Edinburgh, Edinburgh, United Kingdom
    \\\{kahraman.kostas, rabia.yasa.kostas, francisco.zampella, firas.alsehly\}@huawei.com}
}

\maketitle
\makeatletter
\def\ps@IEEEtitlepagestyle{%
  \def\@oddfoot{\mycopyrightnotice}%
  \def\@oddhead{\hbox{}\@IEEEheaderstyle\leftmark\hfil\thepage}\relax
  \def\@evenhead{\@IEEEheaderstyle\thepage\hfil\leftmark\hbox{}}\relax
  \def\@evenfoot{}%
}
\def\mycopyrightnotice{%
  \begin{minipage}{\textwidth}
  \centering \scriptsize
  Copyright~\copyright~2022 IEEE. Personal use of this material is permitted. However, permission to use this material for any other purposes must be obtained from the IEEE by sending a request to pubs-permissions@ieee.org.
  \end{minipage}
}
\makeatother
\maketitle
\begin{abstract}
In recent years WiFi  became the primary source of information to locate a person or device indoor. Collecting RSSI values as reference measurements with known positions, known as WiFi fingerprinting, is commonly used in various positioning methods and algorithms that appear in literature.  However, measuring the spatial distance between given set of WiFi fingerprints is heavily affected by the selection of the signal distance function used to model signal space as geospatial distance. In this study, the authors proposed utilization of machine learning to improve the estimation of geospatial distance between  fingerprints. This research examined data collected from 13 different open datasets  to provide a broad representation aiming for general model that can be used in any indoor environment. The proposed novel approach extracted data features by examining a set of commonly used signal distance metrics via feature selection process that includes feature analysis and genetic algorithm. To demonstrate that the output of this research is venue independent, all models were tested on datasets previously excluded during the training and validation phase. Finally, various machine learning algorithms were compared using wide variety of evaluation metrics including ability to scale out the test bed to real world unsolicited datasets.

\end{abstract}

\begin{IEEEkeywords}
WiFi fingerprinting, RSSI, machine learning, distance estimation indoor positioning, supervised learning
\end{IEEEkeywords}

\section{Introduction}
In the last decade, the proliferation of smart devices has increased significantly, their processing power and capabilities allow users to access services that facilitate their life like communication, wireless payment, messaging, navigation, maps, etc. Many of those services rely on the user's position to adjust the information shown, and the most common way to obtain it outdoor, is through Global Navigation Satellite Systems (GNSS) like GPS, GLONASS, BEIDOU or GALILEO. In indoor environments like offices, shopping malls, subway stations, among others, the common no line of sight to the satellites prevent it from providing an accurate position \cite{rizos2013locata} and the smart devices require additional sources to obtain the location information. 

Many technologies had been used to estimate the position of a user indoors \cite{Mendoza2019}, including ultrasound \cite{ward97}, visible light communication \cite{Do2016}, visual odometry \cite{aqel2016review}, magnetic fields \cite{pasku2017magnetic}, dead reckoning \cite{diaz2019review}, bluetooth \cite{faragher2015location}, Ultra Wide Band \cite{mazhar2017precise}, Wi-Fi \cite{liu2020survey}, among others. This research will focus in the use of Wi-Fi technology and its use in positioning due to its wide availability in all of those indoor environments. A Wireless Access Point (WAP) can provide different measurements to use for positioning, however the Received Signal Strength (RSS) \cite{ding2013overview} has become the most common information source due to its low cost and easy implementation \cite{mainetti2014survey}. This measurement can be used in many different ways to estimate positions, initial works replicated the range estimation of GNSS systems, estimating the distance between a WAP with a known position and the device from the power decay of the signal \cite{koo2010localizing}, however indoors, the multipaths and attenuations alter the propagation in a non predictable way. 

The most common approach to address the non regularity of the power decay with respect to the position is to take samples (fingerprints) of the RSS values from multiple WAPs in a grid of reference points. The position of a new fingerprint can then be estimated from the similarity or distance between the new signal and the previous reference points \cite{khalajmehrabadi2017modern}. The signal differences can use several metrics (Euclidean, Manhattan, Jaccard, Cosine, etc. \cite{torres2015comprehensive}), and from those values, the position can be estimated using deterministic, probabilistic or pattern recognition methods.

It has been observed that the signal difference used affects significantly the performance of the positioning algorithm and the authors in this paper propose to use data science methodologies to improve the estimation of the spatial distance between 2 fingerprints. Most fingerprint and distance estimation methods are based on specific datasets of almost laboratory conditions, the authors propose to use a multi-environment training, and isolated dataset testing as a prediction of the performance of the distance estimation in unknown environments.

The remainder of this paper is structured as follows: in Section \ref{Related Work} related studies in the field are discussed, in Section \ref{Contribution} contribution and how the method serve the Indoor Positioning System (IPS) are described, in Section \ref{Methods} the system architecture and details of the experiments are explained. Section \ref{Results} contains the results of the experiments and  discussion, and in Section \ref{Conclusion} the conclusions are presented. 

\section{Related Work} \label{Related Work}
Recently, many Artificial Intelligence (AI) techniques; such as Machine Learning (ML) and Deep learning (DL) have been used in the solution of indoor positioning problems. The ability of data science to solve high complexity real life problems without the need for exact rules and equations has made AI methods very attractive \cite{nessa2020survey, roy2021survey} for positioning problems. In the IPS field, RADAR\cite{bahl2000radar} was one of the first research works to propose a ML based solution using K-nearest neighbours (KNN) to estimate the location of a user with a few meters error distance. Since then other algorithms were also tested; Gradient Boosted Classifier (GBC) and Random Forest Classifier (RF) in \cite{sapiezynski2017inferring}, Deep Neural Networks (DNN) based classification in\cite{liu2018wifi}, and novelty Graph-based classification algorithm (Hierarchical Navigable Small World Graphs) in\cite{lima2018efficient}.

KNN and its extension weighted K-nearest neighbours (WKNN) have remained quite common for indoor positioning using a reference set, and it is usually referred as fingerprinting. The work of \cite{bahl2000radar} was based on the Euclidean distance between the fingerprints, but as many authors have highlighted, the accuracy of the system is heavily affected by the distance metric used \cite{torres2015comprehensive}, \cite{bahl2000radar}, \cite{farshad2013microscopic}. That distance estimation can be applicable not only for localization, but also tracking of mobile devices and its interactions. This can be valuable for the inference of the network of social relationships, and recently it has helped in the contact tracing of positive cases of COVID-19. In \cite{tu2021epidemic} and \cite{sapiezynski2017inferring}, the authors deduced the proximity network, based on the trajectories and WiFi signal distances between mobile devices. 

Currently, the most common distance (or feature) used for fingerprinting positioning is the Euclidean distance, but subsequent studies improve the metric by extending the feature list with new distances; Jaccard, Manhattan\cite{sapiezynski2017inferring}, Kendall\cite{liu2018wifi}, correlation coefficients\cite{luo2019indoor}, \cite{xie2016improved}, Sørensen and 53 other distances \cite{torres2015comprehensive}. Although, most works measure the quality with one distance or a combination of a few at a time, only \cite{liu2018wifi} tries to estimate the distance using many distance features, training a deep neural network to estimate the real distance from 14 distance features.

These applications are trained and tested on the data obtained from single buildings such as universities~\cite{wei2019calibrating,radu2013himloc} or the well known UJI Indoor Localization dataset \cite{lima2018efficient,torres2015comprehensive} that provides real life data from multiple floors.

\section{Motivation} \label{Motivation}
Today, people spend a lot of time in large buildings such
as hospitals, airports, shopping malls. It is very
important, sometimes even vital, to find directions or locate personnel and equipment in a building. In such areas, where satellite systems do not work efficiently, positioning using WiFi signals and machine learning methods are quite common. However, most of the common studies train machine learning models using data collected only from a single or few collocated buildings. Furthermore, it is widely common to use the same venue for training and testing which render such approach as prone to overfitting and unreliable in unknown environments. To deal with this problem, we propose a generalized model trained as venue independent. We expect that this model will have a wider use with a better extrapolation to unknown scenarios. 

Furthermore, It has been observed that most available research in this concept use a single or limited combination of signal modelling distance functions to estimate the similarity between 2 fingerprints. Given that each distance function usual provide context for a specific characteristics in the signal distribution, which obviously cannot work everywhere, we propose to analyse the influence of each function separately as a feature in our distance modelling. Therefore, expanding the context of one machine learning model to utilize more attributes and functions of available WiFi data has motivated our work in this paper.

\section{Contribution} \label{Contribution}
In this study, we proposed a distance estimation model that detects whether two fingerprints are close to each other and estimates the approximate distance between RSS fingerprints using supervised ML methods; regression and classification. With this study, we have made the following contributions.

\begin{itemize}
	\item We conducted a research with focusing on reproducibility. Our steps are easy to follow to create a new models and applicable on new data from different structures; buildings, train stations, airports etc. 
	
	\item We investigate a broad spectrum of ML and DL algorithms. We assume that it will be a good reference point to comparison for future studies. 
	
	\item Instead of using one dataset obtained from one building, which used frequently for both training and testing purposes in previous studies, we have reserved 2 datasets for testing purposes only, 13 different datasets in total.  We used strictly isolated training and testing data collected from separate buildings. In order to optimize the parameters and estimate distances initially, we used third dataset named validation. In this way, we think that  more generalizable models and realistic results can be obtained. 
	
	\item We do not only estimate the distance between two points (with regression), but also measure the success of our system in predicting the closeness of results (with classifier). 
	
	\item We made the dataset, code used and results obtained publicly available\footnote{Source code available at:\href{https://github.com/kahramankostas/WiFi-Fingerprint}{github.com/kahramankostas/WiFi-Fingerprint}}. 
\end{itemize}

\section{Methodology} \label{Methods}
In this section, dataset, feature extraction, features, ML algorithms and evaluation metrics are explained. 
\subsection{Data}
Many studies uses datasets from the same building for the model training and testing. However, the goal of this paper is to create a generalizable model that can be applicable in any indoor environments. In this respect, while making the dataset that will be used in our study, it was prioritized to have a high representation ability. For this purpose, 11 datasets were used \cite{adriano_moreira_2019_3342526, 
	adriano_moreira_2020_3778646, 
	antonio_ramon_jimenez_ruiz_2016_2791530, 
	antonio_ramon_jimenez_ruiz_2017_2823924, 
	antonio_ramon_jimenez_ruiz_2018_2823964, 
	joaquin_torres_sospedra_2020_4314992, lohan_elena_simona_2017_889798, 
	philipp_richter_2018_1161525, 
	torres2014ujiindoorloc, 
	ipin2019indoor, 
	lohan_elena_simona_2021_5174851} from the \href{http://ipin-conference.org/resources.html}{IPIN conference resources} and additionally two datasets collected by Huawei\footnote{These datasets were presented to the competitors in the 
	\href{https://huawei-uk-challenge.bemyapp.com/}{University Challenge Competition 2021} with the concept of "Data Science for Indoor positioning" organized by Huawei-UK. }. The most important difference of Huawei datasets from other datasets is that the data is collected from shopping centers. Therefore, the new datasets better represent the indoor data samples encountered on daily basis.
\begin{figure}[ht]
	\centerline{\includegraphics[width=1\columnwidth]{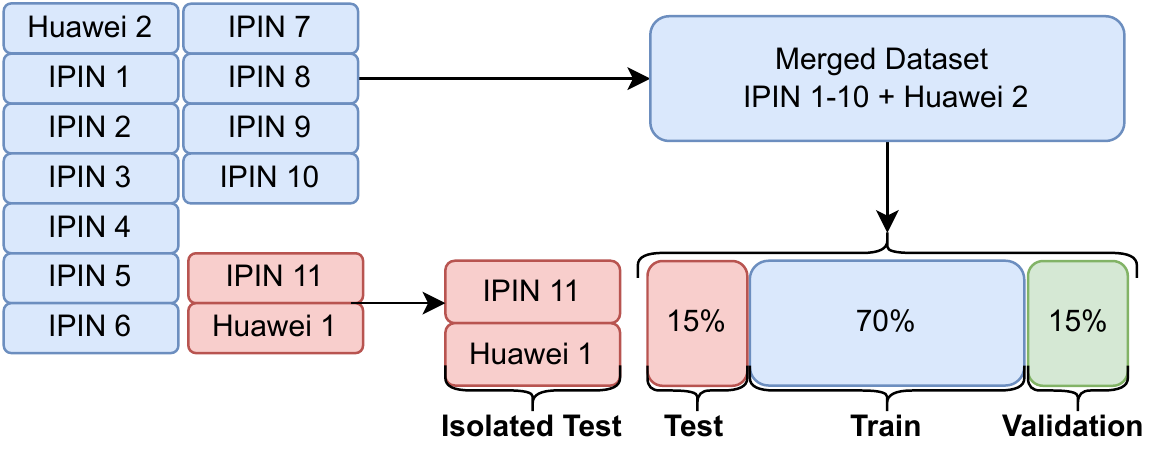}}
	\caption{The creating process of training, validation, testing and isolated test datasets from the data. }
	\label{fig:fiagramofdatasets}
\end{figure}

A large dataset was created by combining the 10 IPIN datasets\cite{adriano_moreira_2019_3342526, 
	adriano_moreira_2020_3778646, 
	antonio_ramon_jimenez_ruiz_2016_2791530, 
	antonio_ramon_jimenez_ruiz_2017_2823924, 
	antonio_ramon_jimenez_ruiz_2018_2823964, 
	joaquin_torres_sospedra_2020_4314992, 
	lohan_elena_simona_2017_889798, 
	philipp_richter_2018_1161525, 
	torres2014ujiindoorloc, 
	ipin2019indoor} and one of the Huawei datasets\cite{huawei2021test}. Thus, we ensured that it contains data from different building structures. The dataset was partitioned into three sets (70\%, 15\%, and 15\%) for training, validation and testing purposes respectively. The validation data was used in the evaluation process of every stages, while retaining the test data for final evaluation independent and isolated. On the other hand, one of the Huawei datasets\cite{huawei2021train} and one of the IPIN datasets\cite{lohan_elena_simona_2021_5174851} were isolated from the initial training, testing and validation data to evaluate the generalization of the system. The creation process of the datasets is also visualized in Fig.~\ref{fig:fiagramofdatasets}.

\subsection{Feature Extraction}

\begin{figure}[ht]
	\centerline{\includegraphics[width=1\columnwidth]{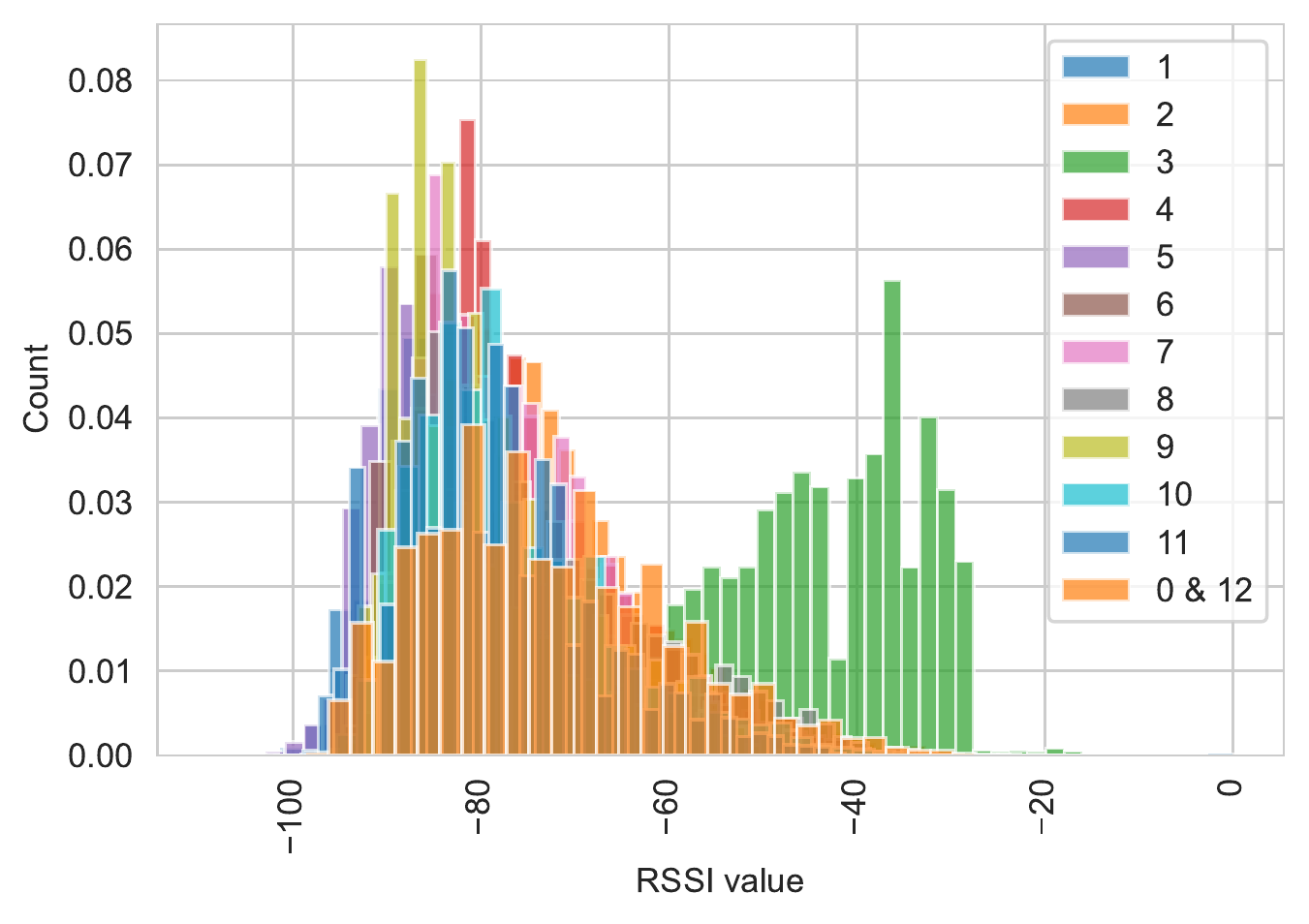}}
	\caption{
		The distribution of RSSI values in all datasets. 
	}
	\label{fig:rssi}
\end{figure}

\begin{figure}[ht]
	\centerline{\includegraphics[width=1\columnwidth]{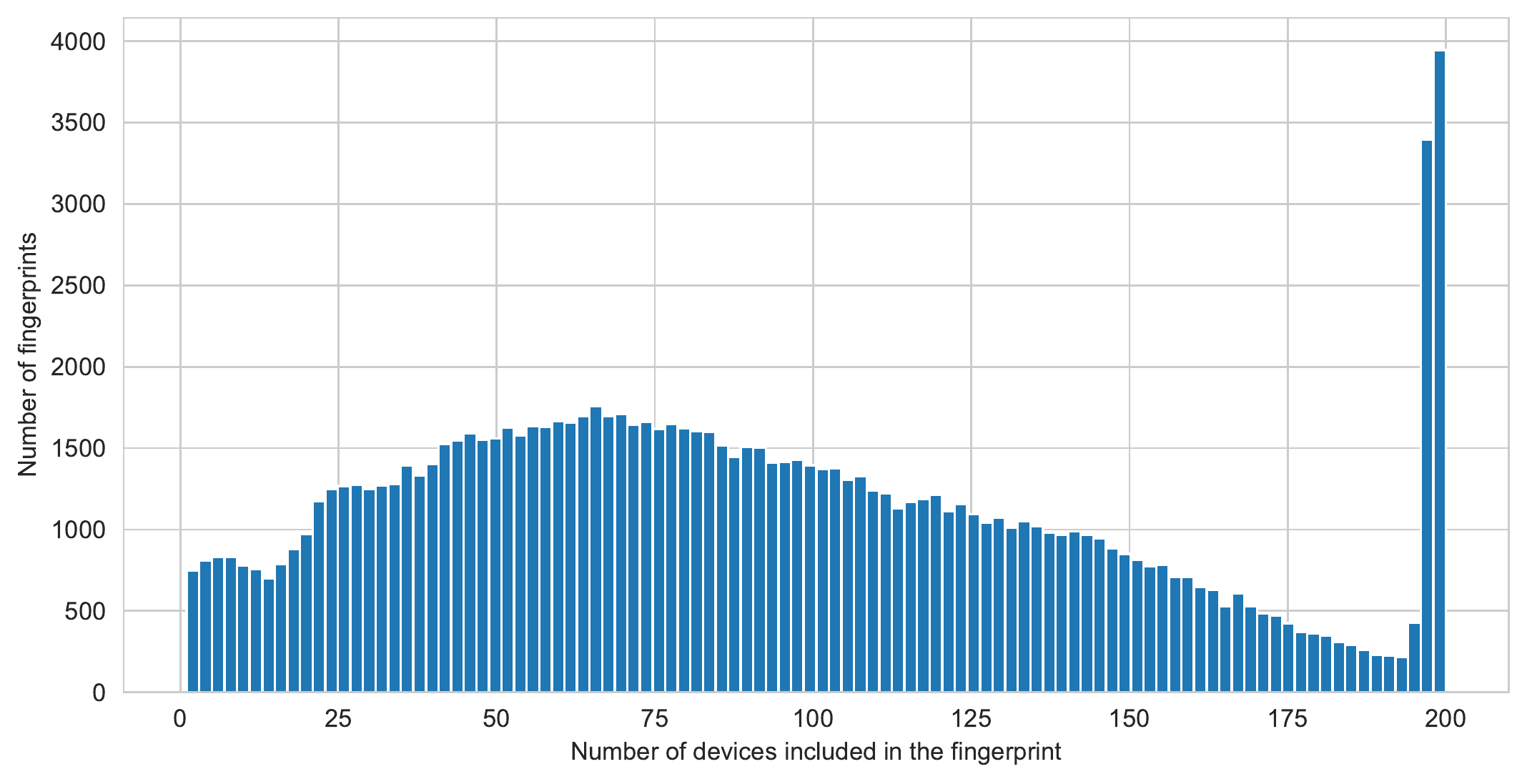}}
	\caption{
		The distribution of fingerprints according to the number of MAC addresses they contain (Huawei datasets\cite{huawei2021test, huawei2021train} )
	}
	\label{fig:number}
\end{figure}

\begin{figure}[ht]
	\centerline{\includegraphics[width=1\columnwidth]{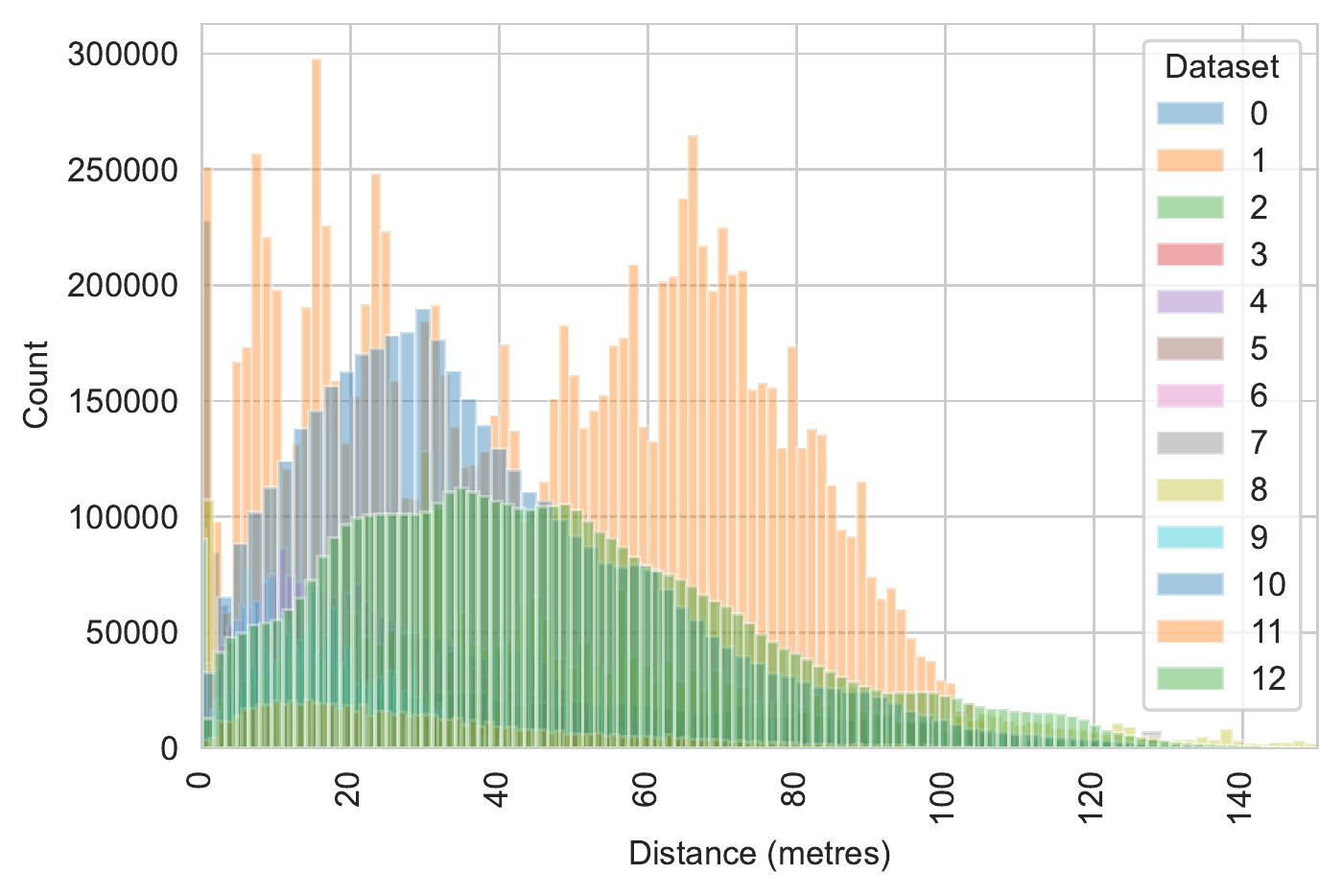}}
	\caption{
		The distribution of distances between fingerprint pairs. 
	}
	\label{fig:distance}
\end{figure}

The distribution of the Received Signal Strength Indicator
(RSSI) values in the datasets is given in Fig.~\ref{fig:rssi}. Almost all
data are tucked between -20 decibel (dB) and -95 dB, except
for some exceptional values in this figure. Ignoring RSSI
values less than -95 dB and greater than -20 dB, outliers are
cleared. In the feature extraction process, RSSI values of the
fingerprints were converted into vectors. The distance between
these vector pairs which representing fingerprints, was calculated with various distance-similarity calculation methods.
The names and formulas of these calculation methods are
given in Table~\ref{tab:denk}. However, the fingerprints in the dataset
are quite different from each other in terms of both the size
and the device they contain. Fig.~\ref{fig:number}  shows the distribution of
fingerprints according to the number of devices they have in
Huawei datasets. While calculating the features, fingerprints
intersected 2 or more Medium Access Control (MAC) addresses were taken into account, the others were excluded. In
addition, only RSSI values of intersecting MAC addresses are
taken into account while calculating the distance, and values
of non-intersecting MAC addresses are ignored. Since datasets do not share any MAC addresses with each other, the distance
between fingerprints in different datasets is not calculated. Two
more features have been added that represent the number of
MAC addresses contained by fingerprint pairs. These are the
number of intersecting and combination MAC addresses. The
actual distance between fingerprints pairs is used as the label
value. The distribution of the distance between the fingerprints
in the datasets is given in Fig.~\ref{fig:distance}.

\begin{table}
	\centering
	\caption{12 distance calculation methods we use in feature
		extraction. $p$: scalar, $w$: weight, $u$  and $v$ : 1-dimensional arrays,
		$m$: point-wise mean of $u$ and $v$, $D$: the \href{http://hanj.cs.illinois.edu/cs412/bk3/KL-divergence.pdf}{Kullback-Leibler divergence}.}
	\begin{tabular}{@{}rl@{}}
		
		\hline
		\rule{0pt}{15pt} Name & Formula \\\hline
		
		\rule{0pt}{15pt}Bray-Curtis & $\frac{\sum{|u_i-v_i|}}{\sum{|u_i+v_i|}}$ \\
		
		\rule{0pt}{15pt} Canberra & $\sum_i \frac{|u_i-v_i|} {|u_i|+|v_i|}$ \\
		
		\rule{0pt}{15pt} Chebyshev & $\max_i {|u_i-v_i|}$ \\
		
		\rule{0pt}{15pt} City Block & $ \sum_i {\left| u_i - v_i \right|}. $ \\
		
		\rule{0pt}{15pt} Correlation & $1 - \frac{ (u - \bar{u}) \cdot (v - \bar{v})}
		{{|| (u - \bar{u})||}_2 {|| (v - \bar{v})||}_2}$ \\
		
		\rule{0pt}{15pt} Cosine & $ 1 - \frac{u \cdot v}
		{||u||_2 ||v||_2}$ \\
		
		\rule{0pt}{15pt} Euclidean & $\sqrt {\sum _{i=1}\left ( u_{i}-v_{i}\right)^2 }$ \\
		
		\rule{0pt}{15pt} Jaccard & $\frac{|u\cap v|}
		{|u| + |v| - |u\cap v|}$ \\

		\rule{0pt}{15pt} Jensen-Shannon & $\sqrt{\frac{D (u \parallel m) + D (v \parallel m)}{2}}$ \\
		
		\rule{0pt}{15pt} Minkowski & $ (\sum{|u_i - v_i|^p})^{1/p}$ \\
		
		\rule{0pt}{15pt} Squared Euclidean & $ {\sum _{i=1}\left ( u_{i}-v_{i}\right)^2 }$ \\
		\rule{0pt}{15pt} Weighted Minkowski & $\left (\sum{ (|w_i (u_i - v_i)|^p)}\right)^{1/p}$ \\ \\
		\hline
	\end{tabular} 
	
	\label{tab:denk}
\end{table}

\subsection{Evaluation Metrics} 

To evaluate the results, the Root mean squared error (RMSE) was used $\sqrt{\frac{1}{n}\Sigma_{i=1}^{n}{({y_i -\hat{y}_i})^2}}$ ($y$ represents the actual value, $\hat{y}$ the estimated value and, $n$ number of samples).

In addition to RMSE, the problem was translated into a classification, using a limit value (4 meters in this case). The value below this limit is accepted as TRUE and the value above it as FALSE, and the regression problem is transformed into binary classification. 
Using this method, the recall can be calculated as $\frac{True Positive (TP)}{TP+False Negative (FN)}$
and the precision as $\frac{TP}{TP+False Positive (FP)}$.  Due to level of noise in the data, our
scoring metric for is biased towards precision over recall. To give a higher weight to the precision the $F_\beta$ score was used (with a $\beta$ value of 0.05), calculated as:
\begin{equation}
F_\beta = \frac{ (1+\beta^2)*Precision*Recall}{\beta^2*Recall+Precision}
\end{equation}

\subsection{Algorithm Selection}
In the datasets, the actual values of the desired outputs along with the fingerprints are shared as the distance in meters between the test points. Based on these data, we can say that this problem is a supervised learning and regression problem. 

Today, deep learning approaches are very successful, especially when data is plentiful like ours. On the other hand, some classical ML methods have shown remarkable success with tabular data~\cite{lundberg2020local2global}. In addition, every dataset is unique, and there is no perfect ML method that succeeds on every dataset. While choosing the ML model, we tried multiple alternative ML methods, considering that each algorithm should be evaluated separately for a dataset. While choosing these ML methods, we tried to provide as wide a scope as possible, but considering that there are hundreds of methods today, it is not practical to try them all, which led us to choose one method from each type of learning. We have shared the ML types and the list of the algorithms we selected from them in Table~\ref{tab:mls}. 
We used keras\ (\href{https://keras.io/}{keras.io/}) for ANN, XGBoost\ (\href{https://github.com/dmlc/xgboost}{github.com/dmlc/xgboost}) for XGBR, and  scikit-learn\ (\href{https://scikit-learn.org/stable/}{scikit-learn.org/stable/}) library for other ML algorithms.

\begin{table}[]
	\caption{The list of ML types and algorithms we used. }
	\centering
	\begin{tabular}{@{}ll@{}}
		\hline
		Machine Learning Type & Machine Learning Algorithm \\
		\hline
		Classical linear regressors & Linear Regression (LR) \\
		
		Tree based & Decision Tree (DTR) \\
		
		Bayesian method based & Bayesian Ridge (BR)\\
		
		Kernel method based & Linear Support Vector (LSVR) \\
		
		Ensemble methods based & eXtreme Gradient Boosting (XGBR) \\
		
		Instance based & K-Nearest Neighbors (KNN) \\
		
		Deep Learning & Artificial Neural Networks (ANN) \\
		\hline
	\end{tabular} 

	\label{tab:mls}
\end{table}

\section{Experiments}
\begin{figure}[ht]
	\centerline{\includegraphics[width=1\columnwidth]{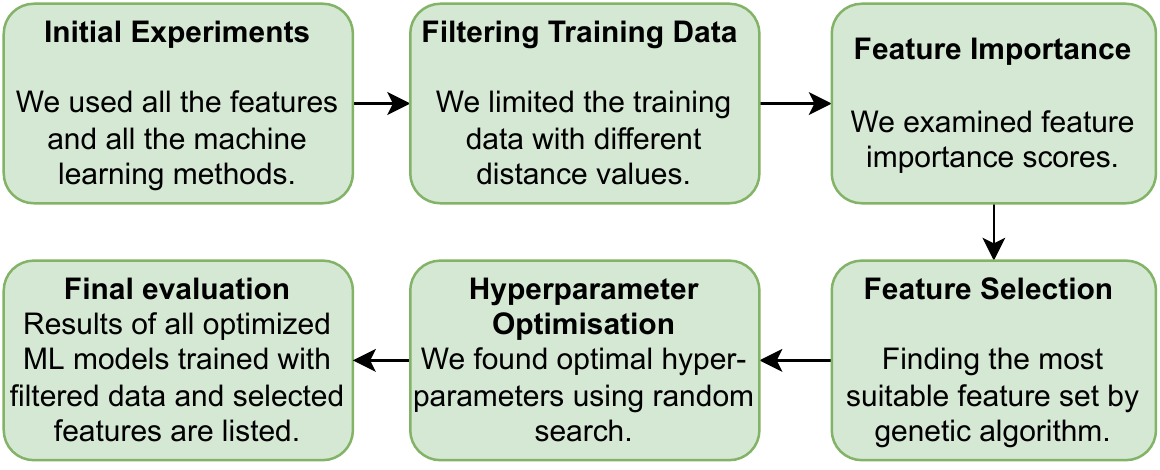}}
	\caption{
		Steps of the experimental process. 
	}
	\label{fig:process}
\end{figure}

Experiments section consists of four subsections. In the first part, an initial evaluation is made using different ML methods with training and validation datasets. In the second part, ideal limits are found by filtering the training data with different thresholds. In the third section, feature selection is made using feature importance scores and genetic algorithm (GA). After that, the ideal hyperparameters are found for each ML algorithm by optimization. In the last part, all ML models are tested with isolated datasets in order to measure the generalizability of the obtained methods. The experiment process is visualized in Fig.~\ref{fig:process}. 
\subsection{Choosing Machine Learning Method}
As an exploratory step, we performed initial experiments using 14 features and the ML methods given in Table~\ref{tab:mls}. 
Table~\ref{tab:resultsone} shows the results obtained from this process. Among ML methods XGB has the highest score followed by ANN and KNN very closely. However, KNN is particularly notable for its extremely high inference time. In real life solutions, it can be a disadvantage. Although LR and NB are two methods with the lowest scores, they stand out with low training and inference time. 

In the next steps, we will observe the results of the LR algorithm in the stages of reshaping our data and feature selecting.
The main reason for selecting LR needing less training and
inference time in these time-consuming processes. In addition,
because it is very simple and deterministic algorithm it does
not contain multiple variables e.g., hyper-parameters, random
seed and gives stable results.

\begin{table}[htbp]
  \centering
  \caption{Comparison of ML algorithms with average of 10 repeats using validation dataset. t is time in seconds. Standard deviation is given with $F_\beta$. The best values are underlined. Hardware: Intel-Core i7-8565U 1.99GHz CPU, 16GB RAM.}
  	\setlength{\tabcolsep}{3pt}
    \begin{tabular}{@{}clrrcrrr@{}}
    \toprule
          & ML    & \multicolumn{1}{l}{Precision} & \multicolumn{1}{l}{Recall} & $F_\beta$  & \multicolumn{1}{l}{RMSE} & \multicolumn{1}{l}{Train-t } & \multicolumn{1}{l}{Test-t} \\
    \midrule
    \multirow{7}[2]{*}{\begin{sideways}Validation\end{sideways}} 
    
        & DTR    & 0.368 & 0.374 & 0.368±0.000 & 23.898 & 1294.1 & 10.421 \\
          & LR    & 0.168 & 0.031 & 0.166±0.000 & 23.345 & 17.798 & 0.225 \\
          & BR    & 0.168 & 0.031 & 0.166±0.000 & 23.345 & \underline{10.646} & \underline{0.201} \\
          & KNNR  & 0.626 & 0.215 & 0.623±0.000 & 21.118 & 637.70& 579.71 \\
          & XGBR  & \underline{0.667} & 0.160 & \underline{0.662±0.000} & \underline{19.271} & 341.44 & 3.180 \\
          & LSVR  & 0.209 & \underline{0.308} & 0.206±0.193 & 110.38 & 17647 & 1.692 \\
          & ANN   & 0.631 & 0.159 & 0.624±0.065 & 19.961 & 6434.8 & 87.941 \\
    \bottomrule
    \end{tabular}%
  \label{tab:resultsone}%
\end{table}%

\subsection{Filtering Training Data}\label{Filtering}

\begin{figure}[ht]
	\centerline{\includegraphics[width=1\columnwidth]{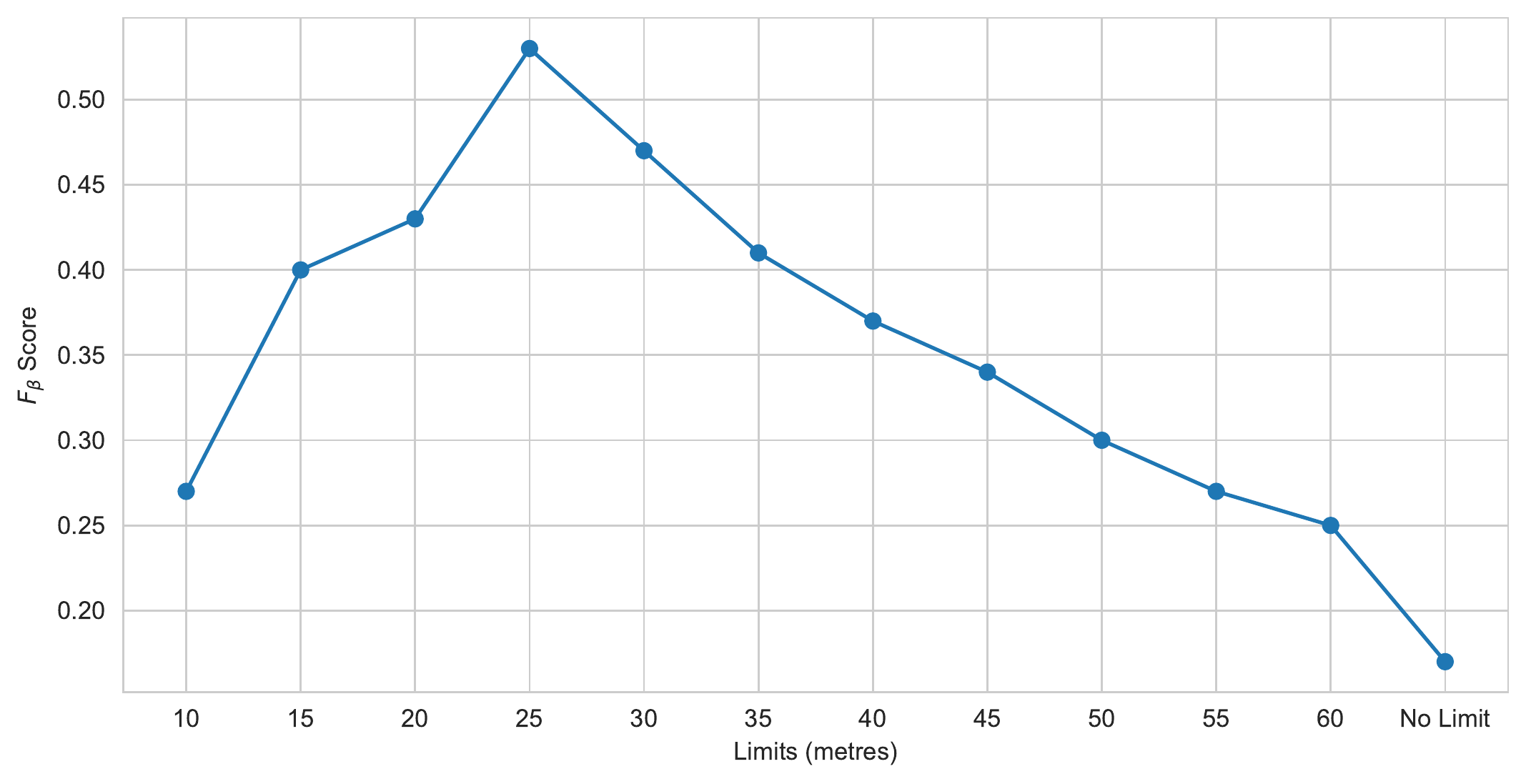}}
	\caption{The change in $F_\beta$ value according to thresholds.}
	\label{fig:fbetta}
\end{figure}

We have shared the distribution of the distances in Fig.~\ref{fig:distance}. According to this figure, the distance between fingerprints varies between 0 and 160 meters and it is quite scattered. In most cases, as the distance between two fingerprints increases, the RSSI decreases and correspondingly, the noise level in the dataset will increase. This noise is likely to affect the model negatively. Therefore, we observed the change of the model by adding various thresholds at training data. For example, a threshold value of 20 means that only fingerprint pairs that are 20 meters or less apart are used in training. In Fig.~\ref{fig:fbetta} the $F_\beta$ results with different thresholds are shared. According to this plot $F_\beta$ increases between 15 and 25 metres, it starts to decrease after 25. With setting the threshold to 25, we increased the $F_\beta$ score from 0.17 to 0.52 for LR. So, we will use the training dataset limited 25 metres which is the peak value, in our next step. 

\subsection{Feature Selection}
\begin{figure}[ht]
	\centerline{\includegraphics[width=1\columnwidth]{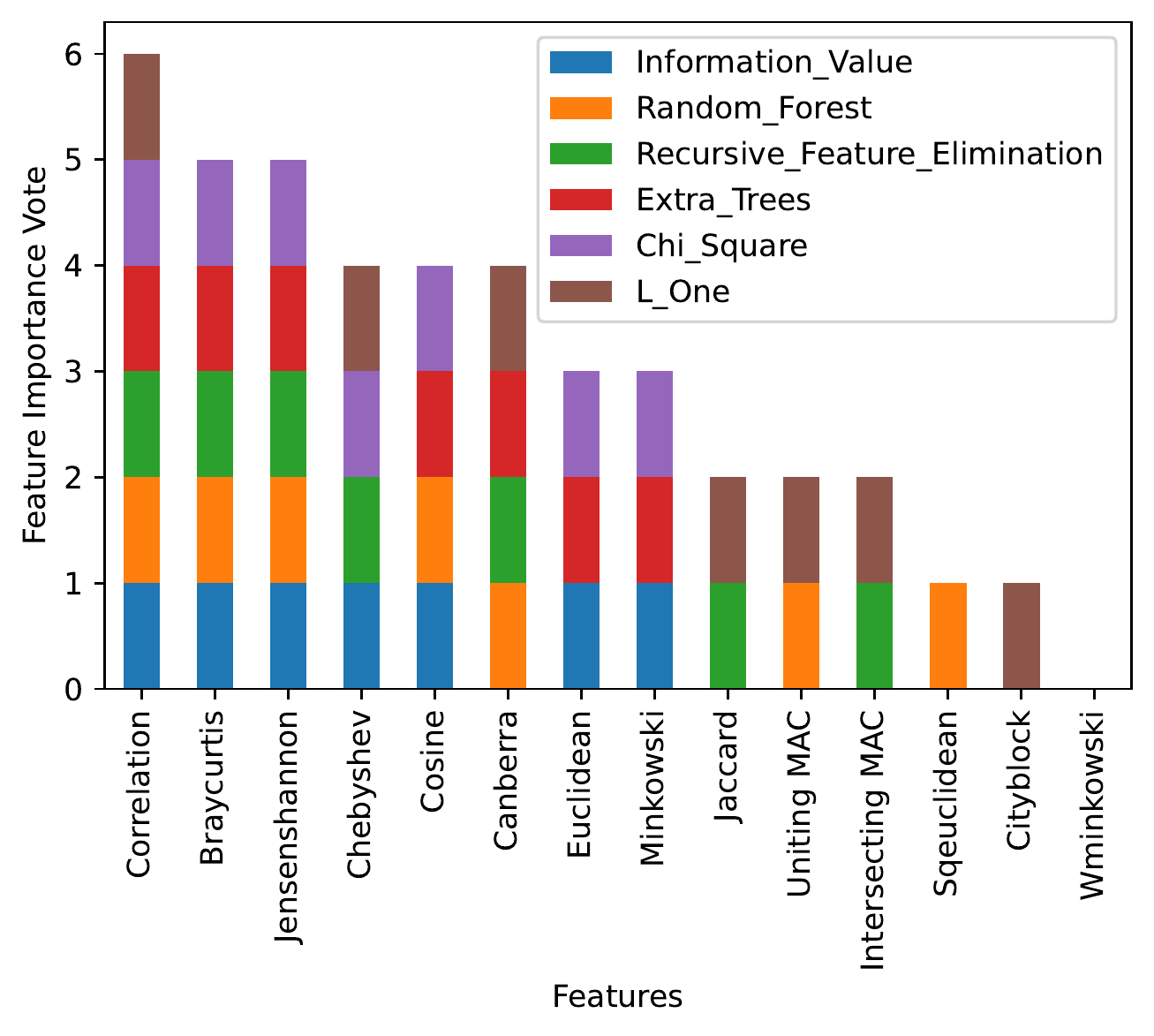}}
	\caption{Features and their votes they receive from feature importance methods.}
	\label{fig:vote}
\end{figure}

\begin{figure}[ht]
	\centerline{\includegraphics[width=1\columnwidth]{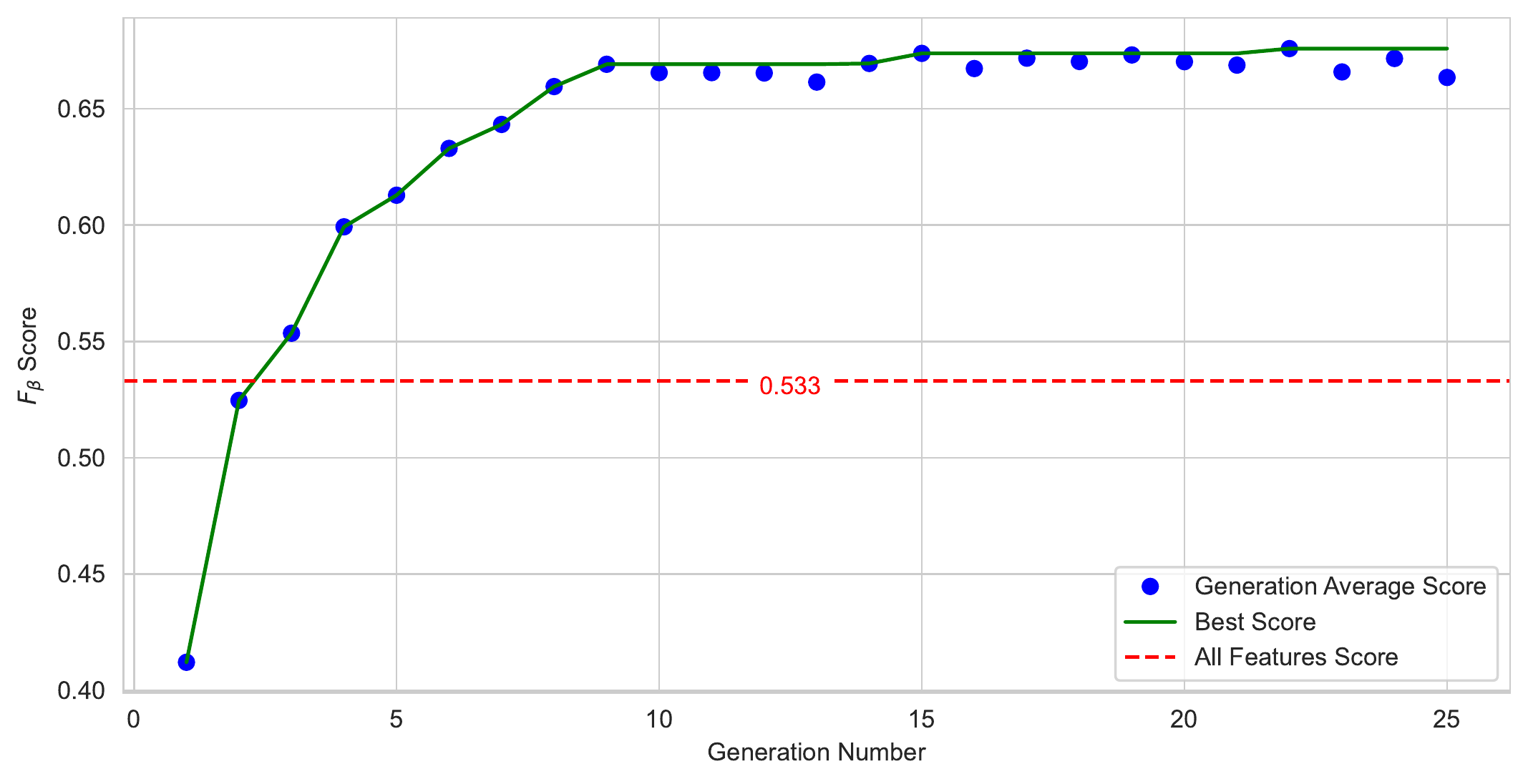}}
	\caption{
		The change in the $F_\beta$ with generations in the GA. 
	}
	\label{fig:GA}
\end{figure}

The model is only as good as the features it uses. That's why we're going to select the most efficient ones among 14 features. 
In the first stage of this process, we will use a voting system\ (\href{https://pypi.org/project/xverse/}{pypi.org/project/xverse/}) to see the importance level of each features. This voting system uses six different feature importance scoring methods (chi-square, extra trees, information value, L1-based, recursive feature elimination, and random forest). Each of these methods votes for a feature as useful or not. The features and the votes they received from each method are given in Fig.~\ref{fig:vote}. 
Although the feature scores give a general idea about the features, they do not allow us to understand which feature group is better. To make up for this point, we used an wrapper method, the GA. It takes $2^{14}$ trials to discover the best feature set of 14 features by trial and error. The GA can create feature sets that perform very well with reasonable computational cost. The $F_\beta$ change occurring parallel to the generations in GA can be seen from the Fig.~\ref{fig:GA}. When all features were used, the score was around 0.53. After GA applied it increased to over 0.65. The GA discarded four out of 14 features and introduced a new feature set with 10 features\footnote{selected features are : Correlation, 
Chebyshev, 
Intersecting MAC, 
Euclidean, 
Cosine, 
Jensenshannon, 
Jaccard, 
Canberra, 
Minkowski, and
Wminkowski}. When we look at the eliminated features importance scores, three of them (Sqeuclidean, Cityblock, and Uniting MAC) can be classified as unimportant, while the other feature (Braycurtis) is classified as quite important. An insignificant feature can become more important if it is combined  with another feature. On the other hand, an important feature may undermine the model by conflicting with another feature or features. Braycurtis feature may have disabled because it conflicts with other features.

After finding the ideal feature set with feature selection, we performed hyper-parameter optimization for the MLs. We used the random search method \ (as used by~\href{https://scikit-learn. org/stable/}{scikit-learn. org/stable/}) for optimization. So, we were able to search for a wide range of parameters within reasonable computational time. In all our processes so far, we have only used the validation dataset to evaluate the performance of steps.

\subsection{Final Experiments} 
So far, we have carried out all experiments and studies using LR. In this step, we adapted all of the experiments to other ML algorithms and reported our final results. For all ML methods, we created models using the training set filtered with a distance of 25 meters and 10 selected features and measured their performance level. In this measurement, in addition to the validation we used at the beginning, we also used other datasets that we isolated.  For this, the model created by using the training dataset was tested on four datasets. After the model training was done once, the same model was used in all four datasets, and the operations were repeated 10 times. The results are given in Table~\ref{tab:resultsall}.
\begin{table}[htbp]
  \centering
  \caption{Comparison of ML algorithms with average of 10 repeats using four datasets. t is time in seconds. Standard deviation is given with $F_\beta$. The best values are underlined.}
  	\setlength{\tabcolsep}{3pt}
    \begin{tabular}{@{}clrrcrrr@{}}
    \toprule
          & ML    & \multicolumn{1}{l}{Precision} & \multicolumn{1}{l}{Recall} & $F_\beta$ & \multicolumn{1}{l}{RMSE} & \multicolumn{1}{l}{Train-t } & \multicolumn{1}{l}{Test-t} \\
    \midrule
    \multirow{7}[2]{*}{\begin{sideways}Validation\end{sideways}} & DTR    & 0.796 & 0.042 & 0.759±0.023 & 35.847 & 27.605 & 1.192 \\
          & LR    & 0.809 & 0.010 & 0.675±0.000 & 36.180 & \underline{3.132}& 0.173 \\
          & BR    & 0.809 & 0.010 & 0.675±0.000 & 36.180 & 4.512 & 0.157 \\
          & KNNR  & 0.606 &\underline{0.104} & 0.599±0.000 & 35.906 & 139.254 & 398.1 \\
          & XGBR  & 0.791 & 0.065 & \underline{0.769±0.000} & 36.191 & 1969.59 & 10.54 \\
          & LSVR  & 0.373 & 0.054 & 0.340±0.244 & \underline{35.453} & 2062.70 & \underline{0.142}\\
          & ANN   & \underline{0.812} & 0.032 & 0.754±0.042 & 35.730 & 4038.79 & 113.5 \\
    \midrule
    \multirow{7}[2]{*}{\begin{sideways}Test\end{sideways}} & DTR    & 0.798 & 0.042 & 0.762±0.025 & 35.847 & 27.605 & 1.166 \\
          & LR    & 0.812 & 0.010 & 0.679±0.000 & 36.178 & \underline{3.132} & 0.158 \\
          & BR    & 0.812 & 0.010 & 0.679±0.000 & 36.178 & 4.512 & 0.157 \\
          & KNNR  & 0.610 & \underline{0.103} & 0.603±0.000 & 35.902 & 139.254 & 399.6 \\
          & XGBR  & 0.794 & 0.065 & \underline{0.773±0.000} & 36.190 & 1969.59 & 10.57 \\
          & LSVR  & 0.374 & 0.055 & 0.343±0.244 & \underline{35.451} & 2062.70 &\underline{0.138} \\
          & ANN   & \underline{0.814} & 0.033 & 0.757±0.038 & 35.730 & 4038.79 & 114.2 \\
    \midrule
    \multirow{7}[2]{*}{\begin{sideways}Huawei\end{sideways}} & DTR    & 0.943 & 0.028 & 0.821±0.091 & 9.384 & 27.605 & 0.183 \\
          & LR    & \underline{1.000} & 0.002 & 0.446±0.000 & 9.833 & \underline{3.132} & 0.033 \\
          & BR    & \underline{1.000} & 0.002 & 0.446±0.000 & 9.833 & 4.512 & 0.028 \\
          & KNNR  & 0.916 & 0.094 & 0.897±0.000 & 9.728 & 139.254 & 53.43 \\
          & XGBR  & 0.970 & 0.034 & \underline{0.907±0.000} & \underline{8.975} & 1969.59 & 1.415 \\
          & LSVR  & 0.788 & \underline{0.219} & 0.746±0.143 & 11.013 & 2062.70 & \underline{0.026}\\
          & ANN   & 0.967 & 0.025 & 0.801±0.140 & 9.217 & 4038.79 & 20.80 \\
    \midrule
    \multirow{7}[2]{*}{\begin{sideways}IPIN\end{sideways}} & DTR    & 0.694 & 0.014 & 0.602±0.083 & 19.473 & 27.605 & 0.091 \\
          & LR    & 0.800 & 0.000 & 0.061±0.000 & 19.921 & \underline{3.132} & 0.017 \\
          & BR    & 0.800 & 0.000 & 0.061±0.000 & 19.921 & 4.512 &\underline{0.013} \\
          & KNNR  & 0.547 & \underline{0.040} & 0.530±0.000 & 19.683 & 139.254 & 33.43 \\
          & XGBR  & \underline{0.891} & 0.031 & \underline{0.833±0.000} & 19.736 & 1969.59 & 0.763 \\
          & LSVR  & 0.689 & 0.036 & 0.444±0.273 & 19.546 & 2062.70 &0.014 \\
          & ANN   & 0.741 & 0.003 & 0.394±0.210 & \underline{19.323} & 4038.79 & 10.22 \\
    \bottomrule
    \end{tabular}%
  \label{tab:resultsall}%
\end{table}%

\section{Results and Discussion} \label{Results}
When we compare the validation part of our final results with the initial experiments, there is a high increase in the success of each of the seven ML algorithms. When we take an overview of Table~\ref{tab:resultsall}, the XGBR method is the most successful method for all datasets, results varying between 0.76 and 0.90, in terms of $F_\beta$ score. It is followed by DT, KNN and ANN methods. If we focus on methods in terms of speed, the XGBR method is reasonably fast, besides its outstanding success. However, the extremely fast and reasonable results of DT may make this method a suitable choice in real life applications. Although LR and BR, which have very short inference times, achieved remarkable success in test and validation datasets, they could not maintain their performance on isolated datasets. The reason for the lower success may be that LR (and similar method BR) is biased against validation and test sets. Since the validation dataset has been used many times with LR in the experimental steps such as feature selection and filtering, this caused an overfitting against the validation and test set (because of both them are created with the training dataset, they have a similar structure). However, the experimental steps contributed to all other ML methods as well.

The proposed models are based on the optimization of a $F_\beta$
score for a classification of fingerprints distances closer than
4 m. However, most of the datasets have uniformly distributed
fingerprints, where the pairwise distances are usually higher
than that value. It has been observed that if all the distance
pairs are allowed in the training, the model will focus more
in the far region, where there is a smaller WiFi overlap and
less correlation with the spatial distance. To minimize this
effect, the training datasets only have fingerprint pairs where
the distance is bellow 25 m, as we have found that increasing
that threshold above that value affects the training, reducing
the $F_\beta$ score and increasing the error in the close region.

\begin{figure}[ht]
	\centerline{\includegraphics[width=1\columnwidth]{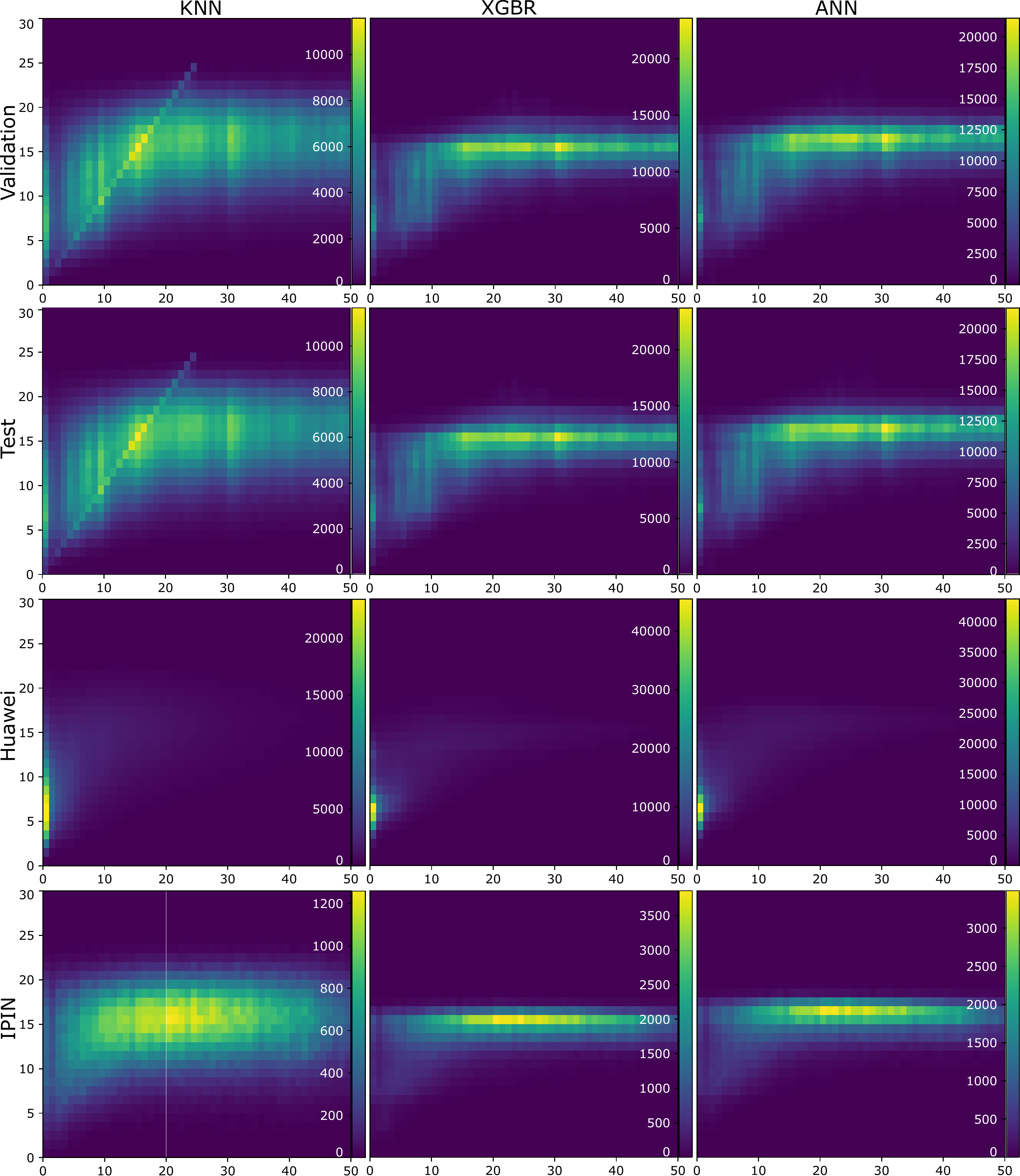}}
	\caption{
		Comparison of real (x-axis) and predicted (y-axis)  data on histogram using three machine learning method and all datasets. Images sorted according to the dataset horizontally and according to the model vertically. 
	}
	\label{fig:hist}
\end{figure}

In addition to the numerical results, the comparison of the estimated and actual values for the three selected models is given in Fig.~\ref{fig:hist}. By examining these figures, a deeper understanding of both the data structure and the distribution of estimated distances can be gained. As can be seen from the figure, especially for distances below 25 meters, the estimates obtained are very compatible with the actual distance. However, there is no estimation of models above this limit. From the figure it can be observed that the KNN model might
have an overfitting problem. After analyzing the data it was
observed that some of the reference points were repeated in several dataset, therefore to have an independent analysis
without overfitting, we will focus on the isolated dataset.

When comparing our results with other studies in the field,
it is difficult to make an exact comparison. There are 2
main reasons for this; (1) Most of the studies use the data
set obtained from the same building for testing and training
purposes. (2) The studies give their results using different
metrics such as; Accuracy, Area under the ROC Curve (AUC),
RMSE etc. We should note that our study mainly focused on
$F_\beta$ value so this may cause lower achievement in other metrics.
While we reported results with $F_\beta$ we also calculated different
metrics for comparison. The NearMe~\cite{krumm2004nearme} study, one of the
pioneering research in the field, carried out the training steps
on the office building dataset and achieved the RMSE of 13.97
meters on the cafeteria building dataset as an unseen test-set.
We achieved RMSE of 8.97 meters with XGBR algorithm on
Huawei dataset, which is a reliable dataset in our experiment
collected from shopping center and reflects the daily life which
was used isolated datasets. Our models gives better results
below 25 metres proximity but our test data demonstrates
skewed distributions (more data above 25 metres), this causes
an increase in the error rate. When we look at the results only
focusing on 25 metres and below, we achieved quite improved
results (see Table~\ref{tab:secondresults}). In the study named Wide~\cite{liu2018wifi}, train and
test sets were splitted from same building data and obtained 3.0 meters in terms of Mean absolute error (MAE) using deep
neural network. Our study achieved MAE of 4.8 metres when
evaluation was made of 25 meters or less between fingerprints
with test data which has similar nature with train data. Also we
got 5.4 metres in Huawei dataset. At first glance, the success
rate seems low compared to the WIDE study, it is a good result
in terms of being a generalizable score when we consider
that our dataset is a composition of 11 different datasets
from various environments. Another study~\cite{sapiezynski2017inferring}, focuses only
proximity inference task between two person and they reported
0.89 in AUC using 10 metres threshold. We achieved 0.907
in terms of $F_\beta$ score on Huawei dataset using XGBR with 4
metres distance.

\begin{table}[htbp]
  \centering
  \caption{Comparison of ML algorithms results for four datasets. Only fingerprint pairs with an actual distance of 25 meters or less between them were evaluated.}
    \begin{tabular}{@{}clrrrrrr@{}}
     \toprule
          & ML    & \multicolumn{1}{l}{MAE} & \multicolumn{1}{l}{RMSE} & \multicolumn{1}{l}{MSE} & \multicolumn{1}{l}{Prec} & \multicolumn{1}{l}{Recall} & \multicolumn{1}{l}{$F_\beta$} \\
    \midrule
    \multirow{7}[2]{*}{\begin{sideways}Validation\end{sideways}} 
          & DTR   & 4.944 & 6.011 & 36.133 & 0.772 & 0.061 & 0.750 \\
          & LR    & 5.343 & 6.378 & 40.684 & \underline{0.823} & 0.010  & 0.684 \\
          & BR    & 5.343 & 6.378 & 40.684 & \underline{0.823} & 0.010  & 0.684 \\
          & KNNR  & 5.065 & 6.362 & 40.475 & 0.662 & \underline{0.104} & 0.653 \\
          & XGBR  & \underline{4.804} & \underline{5.848} & \underline{34.197} & 0.791 & 0.065 &\underline{0.770} \\
          & LSVR  & 6.336 & 7.792 & 60.720 & 0.144 & 0.068 & 0.143 \\
          & ANN   & 4.863 & 5.925 & 35.105 & 0.766 & 0.054 & 0.741 \\
    \midrule
    \multirow{7}[2]{*}{\begin{sideways}Test\end{sideways}} & DTR   & 4.948 & 6.017 & 36.209 & 0.776 & 0.060  & 0.754 \\
          & LR    & 5.346 & 6.384 & 40.757 & \underline{0.831} & 0.010  & 0.692 \\
          & BR    & 5.346 & 6.384 & 40.757 & \underline{0.831} & 0.010  & 0.692 \\
          & KNNR  & 5.064 & 6.365 & 40.508 & 0.665 & \underline{0.103} & 0.656 \\
          & XGBR  & \underline{4.808} & \underline{5.854} & \underline{34.271} & 0.795 & 0.065 & \underline{0.773} \\
          & LSVR  & 6.332 & 7.787 & 60.631 & 0.146 & 0.069 & 0.145 \\
          & ANN   & 4.867 & 5.933 & 35.200 & 0.763 & 0.054 & 0.738 \\
    \midrule
    \multirow{7}[2]{*}{\begin{sideways}Huawei\end{sideways}} & DTR   & 6.244 & 7.103 & 50.453 & 0.951 & 0.009 & 0.756 \\
          & LR    & 6.935 & 7.688 & 59.112 & 1.000     & 0.002 & 0.446 \\
          & BR    & 6.935 & 7.688 & 59.111 & 1.000     & 0.002 & 0.446 \\
          & KNNR  & 6.386 & 7.511 & 56.416 & 0.918 & 0.094 & 0.898 \\
          & XGBR  & 5.454 & \underline{6.284} & \underline{39.489} & \underline{0.970}  & 0.034 & \underline{0.907} \\
          & LSVR  & \underline{5.403} & 6.611 & 43.711 & 0.651 & \underline{0.357} & 0.650 \\
          & ANN   & 5.800 & 6.643 & 44.135 & 0.965 & 0.025 & 0.881 \\
    \midrule
    \multirow{7}[2]{*}{\begin{sideways}IPIN\end{sideways}} & DTR   & 5.441 & 6.526 & 42.588 & 0.717 & 0.029 & 0.677 \\
          & LR    & 5.480  & 6.525 & 42.575 & 1.000& 0.000&0.062 \\
          & BR    & 5.480  & 6.525 & 42.575 & 1.000 & 0.000&0.062 \\
          & KNNR  & 5.783 & 7.059 & 49.823 & 0.669 & 0.040  & 0.643 \\
          & XGBR  & \underline{5.344} & \underline{6.388} & \underline{40.802} & 0.895 & 0.031 & \underline{0.836} \\
          & LSVR  & 5.736 & 6.841 & 46.804 & 0.486 & \underline{0.046} & 0.475 \\
          & ANN   & 5.380  & 6.467 & 41.827 & \underline{0.904} & 0.005 & 0.628 \\
    \bottomrule
    \end{tabular}%
  \label{tab:secondresults}%
\end{table}%

\section{Conclusion} \label{Conclusion}
In our study, we developed a machine learning-based distance estimation model that does not depend on specific
MAC vectors and does not require per venue training, which
detects whether two fingerprints are close to each other and
estimates the approximate distance between fingerprints. For
this process, from the RSSI values we extracted 14 different
features that represent relationships between fingerprint pairs
using various difference metrics. After analyzing the importance score of these features, we found the best performing
feature set using genetic algorithm. During these processes, we
isolated the training and test data from each other to increase
the robustness of the inference and to measure its extrapolation capabilities in unknown environments. The datasets we used
in the final consist of data that it never sees in the training of
the models and the subsequent feature selection and optimization steps. Thus, our results are generalizable and far from
overfitting.

\bibliographystyle{IEEEtran}
\bibliography{references}
\end{document}